\documentclass[10pt,twocolumn,letterpaper]{article}

\usepackage{iccv}
\usepackage{times}
\usepackage{epsfig}

\usepackage[pagebackref=true,breaklinks=true,letterpaper=true,colorlinks,bookmarks=false]{hyperref}

\iccvfinalcopy 

\def\iccvPaperID{10757} 

\ificcvfinal\fi

\usepackage{graphicx}
\usepackage[format=plain,labelformat=simple,labelsep=period,font=small,compatibility=false]{caption}
\usepackage[font=footnotesize,skip=3pt,subrefformat=parens]{subcaption}
\usepackage{makecell}
\usepackage{booktabs}
\usepackage{bm}
\usepackage{xcolor}
\usepackage{float}
\usepackage{multirow}
\usepackage{longtable}
\usepackage{framed}
\usepackage{enumitem}
\usepackage{amsthm,amsmath,amssymb}
\usepackage{pythonhighlight}
\usepackage{appendix}
\usepackage[utf8]{inputenc}
\usepackage{bbding}
\usepackage{tabularx}
\usepackage{dsfont}

\DeclareMathAlphabet{\mathpzc}{OT1}{pzc}{m}{it}

\DeclareMathOperator*{\argmin}{arg\,min}
\usepackage{mathtools}
\usepackage{multirow}
\usepackage{algorithm}
\usepackage[noend]{algpseudocode}
\usepackage{color,soul}

\usepackage{xspace}
\def\Method{GPFL\xspace}
\def\compo{GCE\xspace}
\def\Key{CoV\xspace}

\makeatletter
\DeclareRobustCommand\onedot{\futurelet\@let@token\@onedot}
\def\@onedot{\ifx\@let@token.\else.\null\fi\xspace}
\def\eg{\emph{e.g}\onedot} 
\def\ie{\emph{i.e}\onedot} 
 
\def\etc{\emph{etc}\onedot} \def\vs{\emph{vs}\onedot}

\makeatother

\usepackage[utf8]{inputenc} 
\usepackage[T1]{fontenc}    
\usepackage{url}            
\usepackage{booktabs}       
\usepackage{amsfonts}       
\usepackage{nicefrac}       
\usepackage{microtype}      

\definecolor{blue_}{RGB}{76, 114, 176}
\definecolor{orange_}{RGB}{221, 132, 82}
\definecolor{upload}{RGB}{47, 85, 151}
\definecolor{download}{RGB}{241, 13, 208}
\definecolor{red_}{RGB}{255, 0, 0}
\definecolor{gray_}{RGB}{127, 127, 127}
\definecolor{green_}{RGB}{1, 128, 0}

\theoremstyle{definition}

\usepackage{pifont}
\usepackage[accsupp]{axessibility}

\usepackage[capitalize]{cleveref}
\crefname{section}{Sec.}{Secs.}
\Crefname{section}{Section}{Sections}
\Crefname{table}{Table}{Tables}
\crefname{table}{Tab.}{Tabs.}

\begin{document}

\title{\Method: Simultaneously Learning Global and Personalized Feature Information for Personalized Federated Learning}


\author{
    Jianqing Zhang$^1$, Yang Hua$^2$, Hao Wang$^3$, Tao Song$^1$\\ 
    \vspace{5pt}
    Zhengui Xue$^1$, Ruhui Ma$^1$\thanks{Corresponding author.}, Jian Cao$^1$, Haibing Guan$^1$ \\
    \vspace{5pt}
    $^1$Shanghai Jiao Tong University\quad $^2$Queen's University Belfast\quad $^3$Louisiana State University\\
    {\tt \small \{tsingz, songt333, zhenguixue, ruhuima, cao-jian, hbguan\}@sjtu.edu.cn} \\
    {\tt \small Y.Hua@qub.ac.uk, haowang@lsu.edu}
}

\maketitle

\begin{abstract}
    Federated Learning (FL) is popular for its privacy-preserving and collaborative learning capabilities. Recently, personalized FL (pFL) has received attention for its ability to address statistical heterogeneity and achieve personalization in FL. 
    However, from the perspective of feature extraction, most existing pFL methods only focus on extracting global or personalized feature information during local training, which fails to meet the collaborative learning and personalization goals of pFL. 
    To address this, we propose a new pFL method, named \textbf{\Method}, to simultaneously learn global and personalized feature information on each client. We conduct extensive experiments on six datasets in three statistically heterogeneous settings and show the superiority of \Method over ten state-of-the-art methods regarding effectiveness, scalability, fairness, stability, and privacy. Besides, \Method mitigates overfitting and outperforms the baselines by up to 8.99\% in accuracy. 
\end{abstract}

\section{Introduction}
\label{sec:intro}

To maximize the value of data generated on massive clients while protecting privacy, Federated Learning (FL), an iterative machine learning scheme, comes along with various applications~\cite{kaissis2020secure, hsu2020federated, liu2020fedvision, li2023revisiting, li2023edge}. 
Traditional FL methods focus on collaborative learning and obtaining a reasonable global model. However, in practice, one single global model cannot meet the requirements of every client and performs poorly due to statistical heterogeneity~\cite{kairouz2019advances, alamgir2022federated, tan2022towards}. 


Recently, personalized FL (pFL) has attracted increasing attention in addressing statistical heterogeneity and achieving personalization in FL~\cite{tan2022towards, t2020personalized, Zhang2023fedcp, zhang2022fedala}. 
From the view of each client, it joins FL for additional server information (\eg, global model parameters) to enhance its model and address the data shortage problem. To obtain high-quality server information, each client also has to provide locally learned information for server aggregation. 
Thus, an ideal pFL is one kind of FL with two goals: (1) aggregating information for collaborative learning and (2) training reasonable personalized models. 
On the other hand, since every client is connected to the external environment and shares certain common information, the data present on each client comprises both global and personalized feature information. 


However, from a feature extraction perspective, existing pFL methods only focus on one of these two goals on clients. For collaborative learning, FedRoD~\cite{chen2021bridging} trains the feature extractor to extract global feature information for its global objective, but it does not extract personalized feature information for personalized tasks. For personalization, FedPer~\cite{arivazhagan2019federated} and FedRep~\cite{collins2021exploiting} only use local data to train the model for the personalized objective, losing some global information during local training~\cite{chen2022metafed}, which is not beneficial for collaborative learning. 
Although FedPHP~\cite{li2021fedphp}/ FedProto~\cite{tan2022fedproto} utilizes global features/prototypes to guide personalized feature extraction, the quality of global features/prototypes depends on the quality of feature extractors, which is paradoxical. Poor global features/prototypes mislead feature extraction in turn. 

To simultaneously learn global and personalized feature information on each client, we propose a novel pFL framework, named \textbf{\Method}.
Inspired by the category anchors that introduce extra common information in domain adaptation~\cite{zhang2019category}, we learn the global feature information with the guidance of global category embeddings using the \textit{Global Category Embedding layer} (${\rm \compo}$). Besides, we learn personalized feature information through personalized tasks. However, learning two contrary (global \vs personalized) objectives is confusing, so we devise and insert the \textit{Conditional Valve} (${\rm \Key}$) after the feature extractor to create a global guidance route and a personalized task route in the client model. With ${\rm \Key}$, we learn global and personalized feature information separately at the same time, unlike FedRoD, FedPer, and FedRep, which only learn one kind of feature information. Besides, \Method leverages trainable category embeddings to guide feature extraction at both the magnitude and angle levels, unlike FedPHP and FedProto, which rely on the well-trained feature extractor. 
Furthermore, the global category embeddings in \Method introduce extra global information besides local data, which can mitigate the overfitting of personalized models and enhance fairness and privacy-preserving ability. 

To evaluate \Method regarding effectiveness, scalability, fairness, stability, and privacy, we compare \Method with ten state-of-the-art (SOTA) methods on six datasets in \textit{Computer Vision} (CV), \textit{Natural Language Processing} (NLP), and \textit{Internet of Things} (IoT) domains. 
Besides, we consider the \textit{label skew}~\cite{mcmahan2017communication, NEURIPS2020_18df51b9, li2022federated},  \textit{feature shift}~\cite{li2020fedbn}, and \textit{real world}~\cite{zhang2022federated, gao2020end} settings to simulate different kinds of statistical heterogeneity in FL. 
Experimental results show that \Method outperforms these baselines by up to 8.99\% in accuracy. We provide the code in the supplementary materials. 
Overall, our key contributions are
\begin{itemize}
\setlength\itemsep{0pt}
    \item We emphasize the importance of achieving both collaborative learning and individualized goals in pFL and propose a pFL method \Method that simultaneously learns the global and personalized feature information. 
    \item We learn the global feature information through trainable category embeddings, and the additional global information in ${\rm \compo}$ mitigates the overfitting of the personalized model to local data. 
    \item We conduct extensive experiments in the CV, NLP, and IoT domains under \textit{label skew}, \textit{feature shift}, and \textit{real world} settings. The results show that our \Method outperforms the SOTA method in terms of effectiveness, scalability, fairness, stability, and privacy. 
\end{itemize}

\section{Related Work \& Background}
\label{sec:related}

\subsection{Personalized Federated Learning}


\noindent\textbf{Meta-learning \& fine-tuning. \ } Per-FedAvg~\cite{NEURIPS2020_24389bfe} and FedMeta~\cite{chen2018federated} are similar methods that learn a global model with the aggregated model update trend to achieve good performance on each client with a few steps of local fine-tuning. However, the aggregated trend cannot meet the model update trends of every client. 

\noindent\textbf{Personalized heads. \ } FedPer~\cite{arivazhagan2019federated}, FedRep~\cite{collins2021exploiting}, and FedRoD~\cite{chen2021bridging} split the given backbone into a feature extractor and a head. FedPer and FedRep are similar methods that only share the feature extractor between the server and clients. 
Different from them, each client in FedRoD owns a feature extractor and two heads. FedRoD trains the feature extractor as well as the shared head for its global objective (with the balanced softmax (BSM) loss~\cite{ren2020balanced}) and trains the personalized head for its personalized objective.
However, it does not derive gradients, w.r.t. the feature extractor from the personalized objective, ignoring personalized feature extraction. Besides, the BSM loss is ineffective in \textit{feature shift} settings. 

\noindent\textbf{Regularization. \ } Different from FedProx~\cite{MLSYS2020_38af8613}, which regularizes the difference between local model parameters and frozen global parameters, pFedMe~\cite{t2020personalized}/Ditto~\cite{li2021ditto} uses a proximal term for the additional personalized models. The personalized model in Ditto benefits from the global parameter guidance. 
However, they ignore the global and personalized feature information extraction. 

\noindent\textbf{Feature extraction guidance. \ } 
FedPHP aligns the features outputted by the personalized feature extractor and the global feature extractor for each sample, and FedProto aligns the feature vectors to their corresponding category prototypes. 
However, FedProto only guides a feature vector to be close to its corresponding prototype rather than guiding it to stay away from other prototypes, which results in the intersection of classification boundaries (see \cref{sec:feat}). 
Besides, FedProto generates prototypes based on the learned feature vectors, so the prototypes in FedProto can be uninformative without the well-trained feature extractor. 
This problem exacerbates when large backbones are trained from scratch in FL because they have difficulty learning good feature extractors in early iterations. 

\subsection{Conditional Computation}

Usually, the structures of most DNNs are static during training and inference. With conditional computing techniques~\cite{bengio2013estimating, garnelo2018conditional, guo2019spottune, oreshkin2018tadam}, such as dynamic routing~\cite{guo2019spottune, liu2018dynamic, wu2018blockdrop}, a DNN can have a dynamic structure when given different conditional inputs. For example, using an auxiliary policy network, SpotTune~\cite{guo2019spottune} dynamically chooses which and how many blocks in a pre-trained residual network should be fine-tuned according to the input images. During inference, BlockDrop~\cite{wu2018blockdrop} executes specific layers of a residual network according to the decisions made by reinforcement learning. 
D$^2$NN~\cite{liu2018dynamic} proposes a kind of DNN that allows selective execution through controller modules. 

These methods are designed in central learning scenarios for specific tasks. Inspired by them, we propose a ${\rm \Key}$ to create global and personalized routes in the client model for global and personalized feature information extraction. 




\section{\Method}

\subsection{Problem Statement}

We have a total of $N$ clients and client $i$ ($i \in [N]$) generates its private data ${\bm x}_i$ (labelled by $y_i$) via a distinct distribution $\mathcal{D}_i$. For the personalized task, we denote the personalized objective on client $i$ as 
\begin{equation}
    \mathcal{F}_i := \mathbb{E}_{({\bm x}_i, y_i) \sim \mathcal{D}_i} \mathcal{L}_i({\bm x}_i, y_i; W_i), \label{eq:per}
\end{equation}
where $\mathcal{L}_i$ is the personalized loss function, and $W_i$ is the parameters of all modules on client $i$. 
For all clients, our objective is 
\begin{equation}
    \{W_1, \ldots, W_N\} = \argmin \ \mathcal{G}(\mathcal{F}_1, \ldots, \mathcal{F}_N), \label{eq:obj}
\end{equation}
where typically $\mathcal{G}(\mathcal{F}_1, \ldots, \mathcal{F}_N) = \sum_{i\in [N]} n_i \mathcal{F}_i$,  $n_i = \frac{|\mathcal{D}_i|}{\sum_{j\in [N]} |\mathcal{D}_j|}$ measures the importance of client $i$ and $|\mathcal{D}_i|$ is the number of training samples on client $i$. With \cref{eq:per}, \cref{eq:obj} considers collaborative learning and personalization. 

\subsection{Method}
\label{sec:method}

\begin{figure}[ht]
	\centering
	\includegraphics[width=\linewidth]{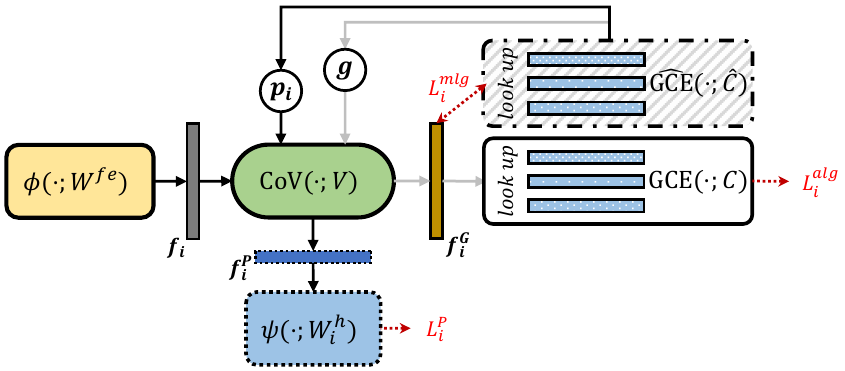}
\setlength{\abovecaptionskip}{-5pt}
\setlength{\belowcaptionskip}{-10pt}
	\caption{Illustration of client modules and data flow between them. Client $i$ shares $W^{fe}$, $V$, $C$, and $\hat{C}$ while keeping $W^{h}_i$ locally. Global category embeddings in $\widehat{\rm \compo}$ are frozen before local training. We simultaneously train $\phi$, ${\rm \Key}$, ${\rm \compo}$, and $\psi$ in an end-to-end manner on the client. For training, we activate both the global guidance route (\textcolor{gray_}{gray arrows}) and the personalized task route (black arrows). For inference and evaluation, only the personalized task route is activated. }
	\label{fig:main}
\end{figure}

\noindent\textbf{Overview. \ } 
Focusing on the extracted features, we follow FedRep and split the backbone into a feature extractor ($\phi$) and a head ($\psi$), where $\psi$ is the last fully connected (FC) layers in the backbone, and $\phi$ represents the remaining layers. In \Cref{fig:main}, to achieve the collaborative learning and personalization goals in pFL, we use ${\rm \Key}$ to transform the original feature vector ${\bm f}_i$ to two feature vectors ${\bm f}_i^G$ and ${\bm f}_i^P$ for global and personalized feature extraction, respectively. Then we learn the global feature information with the guidance of global category embeddings in ${\rm \compo}$ and learn personalized feature information through personalized tasks. 
Using the embedding technique~\cite{vaswani2017attention, zhang2021tlsan}, we can get any trainable category embedding ${\bm u}$ through the $look\ up$ operation: ${\bm u} = {\rm \compo}(u; C)$, where $u$ is a category ID. 
In each iteration, we share global parameters $W^{fe}$, $V$, and $C$ among clients and obtain $\hat{C}$ by copying $C$ after receiving $C$. 
We denote the trainable parameters as $W_i := \{W^{fe}, V, C, W_i ^h\}$. 

\noindent\textbf{Feature extraction. \ } We firstly obtain ${\bm f}_i$ through $\phi$ that maps the data samples to a lower feature space of dimension $K$: $\phi: \mathbb{R}^D \rightarrow \mathbb{R}^K$, where typically $D \gg K$. Formally, $\forall ({\bm x}_i, y_i) \sim \mathcal{D}_i, {\bm f}_i = \phi({\bm x}_i; W^{fe})$.

\noindent\textbf{Feature transformation. \ } Inspired by the conditional computation techniques~\cite{guo2019spottune, hua2021temporal, oreshkin2018tadam}, we transform ${\bm f}_i$ to ${\bm f}_i^G$ and ${\bm f}_i^P$ through the affine mapping~\cite{bengio2013representation, hua2021temporal, oreshkin2018tadam}:
\begin{equation}
\begin{aligned}
    {\bm f}_i^G &= \sigma[({\bm \gamma} + {\bm 1}) \odot {\bm f}_i + {\bm \beta}],\\
    {\bm f}_i^P &= \sigma[({\bm \gamma}_i + {\bm 1}) \odot {\bm f}_i + {\bm \beta}_i], 
\end{aligned}
\end{equation}
where ${\bm 1}$ has the same shape as ${\bm f}_i$ with all the values equal to 1 and $\sigma$ is the ReLU activation function~\cite{li2017convergence}. $\odot$ is the Hadamard product. ${\bm \gamma}$, ${\bm \beta}$, ${\bm \gamma}_i$, and ${\bm \beta}_i$ are generated by ${\rm \Key}$:
\begin{equation}
\begin{aligned}
    \{{\bm \gamma}, {\bm \beta}\} &= {\rm \Key}({\bm f}_i, {\bm g}; V),\\
    \{{\bm \gamma}_i, {\bm \beta}_i\} &= {\rm \Key}({\bm f}_i, {\bm p}_i; V), \label{eq:gamma_gen}
\end{aligned}
\end{equation}
where ${\bm g} \in \mathbb{R}^K$ and ${\bm p}_i \in \mathbb{R}^K$ are the global and personalized conditional input (described later), respectively. ${\rm \Key}$ consists of two sub-modules ${\rm \Key}_\gamma$ and ${\rm \Key}_\beta$ with identical structures but different parameters. Concretely, ${\rm \Key}_\gamma$ generates ${\bm \gamma}$/${\bm \gamma}_i$ by inputting ${\bm g}$/${\bm p}_i$ sequentially to an FC layer, a ReLU activation, and a layer-normalization layer~\cite{ba2016layer}. ${\rm \Key}_\beta$ generates ${\bm \beta}$/${\bm \beta}_i$ in a similar way. 

\noindent\textbf{Generating ${\bm g}$ and ${\bm p}_i$. \ } ${\bm g}$ is identical among clients, while ${\bm p}_i$ contains local data distribution information. Since all the clients share the same $\hat{C}$ during local training, we can generate ${\bm g}$ by averaging all the frozen category embeddings (with dimension $K$):
\begin{equation}
    {\bm g} = \frac{\sum_{u\in [U]} \widehat{\rm \compo}(u; \hat{C})}{U}. \label{eq:gen_g}
\end{equation}
For ${\bm p}_i$, we first obtain data distribution information through the statistics on client $i$. Specifically, we generate the proportion coefficient of category $u$ by 
\begin{equation}
    \alpha_i^u = \mathbb{E}_{({\bm x}_i, y_i) \sim \mathcal{D}_i} \mathbb{I}\{y_i = u\}, u \in [U],
\end{equation}
where $\mathbb{I}\{\cdot\}$ is the indicator function, and $U$ is the number of all the categories among clients. Then, we have 
\begin{equation}
    {\bm p}_i = \frac{\sum_{u\in [U]} \widehat{\rm \compo}(u; \hat{C}) \cdot \alpha_i^u}{U}. \label{eq:gen_p}
\end{equation}

\noindent\textbf{Angle-level global guidance. \ } In addition to sharing model parameters between server and clients, sharing other information, such as prototype ~\cite{tan2022fedproto} and logits~\cite{zhang2021parameterized}, has shown effectiveness in the literature. Thus, we propose to share the global category embeddings across clients. Inspired by the contrastive loss~\cite{li2021model, chen2020simple, sohn2016improved}, we guide each feature vector to be close to its corresponding category embedding while staying away from other category embeddings, which spreads out the category embeddings during training. Formally, we have the angle-level guidance loss
\begin{equation}
    \mathcal{L}_i^{alg} = - \log \frac{\exp{({\rm sim}({\bm f}_i^G, {\rm \compo}(y_i; C)))}}{\sum_{u\in [U]} \exp{({\rm sim}({\bm f}_i^G, {\rm \compo}(u; C)))}}, \label{eq:la}
\end{equation}
where ${\rm sim}({\bm f}, {\bm v}) = \frac{{\bm f}^T {\bm v}}{||{\bm f}||_2 ||{\bm v}||_2}$ is a cosine similarity function. 
Due to the L2-norm $||\cdot||_2$, ${\rm sim}(\cdot)$ only measures the similarity between ${\bm f}_i^G$ and ${\rm \compo}(u; C), u \in [U]$ at the angle level ignoring their magnitude. Note that, all the embeddings in ${\rm \compo}$ are updated through \cref{eq:la}. 

\noindent\textbf{Magnitude-level global guidance. \ } Inspired by the proximal term that keeps the local model parameters close to the frozen global parameters~\cite{MLSYS2020_38af8613}, we keep ${\bm f}_i^G$ close to its corresponding frozen global embedding. Formally, we have the magnitude-level guidance loss
\begin{equation}
    \mathcal{L}_i^{mlg} = ||{\bm f}_i^G, \widehat{\rm \compo}(y_i; \hat{C})||_2. 
\end{equation}

\noindent\textbf{Personalized tasks. \ } With ${\bm f}_i^P$, client $i$ learns a head $\psi$ that maps from the transformed feature space to the category space: $\psi: \mathbb{R}^K \rightarrow \mathbb{R}^U$. Formally, we have 
\begin{equation}
    \mathcal{L}_i^P = \ell(\psi({\bm f}_i^P; W_i^h), y_i), 
\end{equation}
where $\ell$ is the \textit{Cross Entropy} (CE) loss function~\cite{zhang2018generalized}. 

\noindent\textbf{Local loss function $\mathcal{L}_i$. \ } Combining the loss mentioned above together, we have 
\begin{equation}
    \mathcal{L}_i = \mathcal{L}_i^P + \mathcal{L}_i^{alg} + \lambda \mathcal{L}_i^{mlg} + \mu ||V||_2 + \mu ||C||_2, 
\end{equation}
where $\lambda$ and $\mu$ are hyperparameters. The entire learning process is shown in \Cref{algo}.

\begin{algorithm}[t]
	\caption{The Learning Process in \Method}
	\begin{algorithmic}[1]
		\Require 
		$N$ clients with their local data; 
		initial parameters $W^{fe, 0}$, $V^{0}$, $C^{0}$; 
		$\eta$: local learning rate; 
		$\lambda$ and $\mu$: hyperparameters;
		$\rho$: client joining ratio; 
		$T$: total training iterations. 
		\Ensure 
		Personalized model parameters $\{W_1, \ldots, W_N\}$.
		\State Client $i, \forall i \in [N]$ initializes its $\psi$ and obtains $W_i^{h, 0}$. 
		\For{iteration $t=0, \ldots, T$}
		    \State Server samples a client subset $\mathcal{I}^t$ based on $\rho$.
		    \State Server sends $\{W^{fe, t}, V^{t}, C^{t}\}$ to $\mathcal{I}^t$.
		    \For{Client $i \in \mathcal{I}^t$ in parallel}
		        \Statex \Comment{\textbf{local initialization}}
		        \State Initialize $\phi$, ${\rm \Key}$, ${\rm \compo}$ with $\{W^{fe, t}, V^{t}, C^{t}\}$.
		        \State Initialize $\widehat{\rm \compo}$ with $C^{t}$. 
		        \State Generate ${\bm g}$ and ${\bm p}_i$ by \cref{eq:gen_g} and \cref{eq:gen_p}.
		        \Statex \Comment{\textbf{local training}}
		        \State Update $W^{fe, t}$, $V^{t}$, $C^{t}$, $W_i^{h, t}$ simultaneously:
		            \State \qquad $W_i^{fe, t} \leftarrow W^{fe, t} - \eta \nabla_{W^{fe, t}} \mathcal{F}_i$; 
		            \State \qquad $V_i^{t} \leftarrow V^{t} - \eta \nabla_{V^{t}} \mathcal{F}_i$; 
		            \State \qquad $C_i^{t} \leftarrow C^{t} - \eta \nabla_{C^{t}} \mathcal{F}_i$;
		            \State \qquad $W_i^{h, t+1} \leftarrow W_i^{h, t} - \eta \nabla_{W_i^{h, t}} \mathcal{F}_i$.
		        \State Upload $\{W_i^{fe, t}, V_i^{t}, C_i^{t}\}$ to the server. 
		    \EndFor
		    \Statex \Comment{\textbf{Server aggregation}}
		    \State Server calculates $n^t = \sum_{i \in
		    \mathcal{I}^t} n_i$ and obtains 
		        \State \qquad $W^{fe, t+1} = \sum_{i \in \mathcal{I}^t} \frac{n_i}{n^t} W_i^{fe, t}$; 
		        \State \qquad $V^{t+1} = \sum_{i \in \mathcal{I}^t} \frac{n_i}{n^t} V_i^{t}$; 
		        \State \qquad $C^{t+1} = \sum_{i \in \mathcal{I}^t} \frac{n_i}{n^t} C_i^{t}$.
		\EndFor
		\\
		\Return $\{W_1, \ldots, W_N\}$
	\end{algorithmic}
	\label{algo}
\end{algorithm}

\subsection{Privacy Analysis}
\label{sec:privacy_ana}

Following FedCG~\cite{wufedcg}, we consider a semi-honest scenario where the server follows the FL protocol but may recover original data from a victim client $i$ with its model updates through \textit{Deep Leakage from Gradients} (DLG)~\cite{zhu2019deep}. Given model (or feature extractor) $\Phi$ and model updates $\Delta$, DLG learns dummy input $\tilde{{\bm x}}$ and dummy logits (or feature vectors) output $\tilde{{\bm z}}$ by minimizing
\begin{equation}
    \mathcal{L}^{dlg} = ||\nabla \tilde{\ell} (\Phi(\tilde{{\bm x}}), \tilde{{\bm z}}) - \Delta||^2, 
\end{equation}
where we use \textit{Mean Squared Error} (MSE) loss~\cite{yoo2019learning} for $\tilde{\ell}$. Then the server can obtain the recovered input $\tilde{{\bm x}}^*$. 

In \Method, $W_i^h$ and $\alpha_i^u, u\in [U]$ are not shared outside the client, protecting most of the private information. 
Without $\alpha_i^u$, the ${\rm \Key}$ module can only perform transformation for the global route, like a regular layer in a backbone. The server can treat the combination of $\phi$ and ${\rm \Key}$ as a pseudo feature extractor $\tilde{\phi} := \phi \ \circ \ {\rm \Key}$ and obtain $\tilde{{\bm f}}^G := \tilde{\phi}(\tilde{{\bm x}}; W_i^{fe, t}, V_i^{t})$. Compared to FedPer and FedRep which only share the feature extractor, either $\phi$ or $\tilde{\phi}$ learns more global information. 
Besides, the server can utilize ${\rm \compo}$ to calculate the logit output for category $u$ by
\begin{equation}
    {\rm logit}_u = {\rm sim}(\tilde{{\bm f}}^G, {\rm \compo}(u; C_i^{t})), u\in [U].
\end{equation}
In other words, the server can use ${\rm \compo}$ by integrating it with $\tilde{\phi}$ as a pseudo model like the uploaded client model in FedAvg and Ditto, but with more global information. 
According to previous work~\cite{nasr2018comprehensive}, the model with more global information has a better privacy-preserving ability. We show the experimental results in \cref{sec:privacy}.

\section{Performance Comparison}
\label{sec:perform}

We evaluate the performance of \Method in terms of the learned features, effectiveness, scalability, fairness, stability, and privacy. Specifically, we compare \Method with ten SOTA methods, including FedAvg~\cite{mcmahan2017communication},  FedProx~\cite{MLSYS2020_38af8613}, Per-FedAvg~\cite{NEURIPS2020_24389bfe}, 
pFedMe~\cite{t2020personalized}, Ditto~\cite{li2021ditto},  FedPer\cite{arivazhagan2019federated}, FedRep~\cite{collins2021exploiting}, FedRoD~\cite{chen2021bridging}, FedPHP~\cite{li2021fedphp}, and FedProto~\cite{tan2022fedproto}, on CV, NLP, and IoT tasks. 

\subsection{Setup} 

\noindent\textbf{Datasets. \ } For CV tasks, we use three public datasets: Fashion-MNIST (FMNIST)~\cite{xiao2017fashion}, Cifar100~\cite{krizhevsky2009learning}, and Tiny-ImageNet~\cite{chrabaszcz2017downsampled}. For NLP tasks, we use two public datasets: AG News~\cite{zhang2015character} and Amazon Review~\cite{feng2021kd3a}. For the IoT task, we use a \textit{Human Activity Recognition} (HAR) dataset~\cite{anguita2012human}. 

\noindent\textbf{Backbones. \ } Following work~\cite{mcmahan2017communication, luo2021no, geiping2020inverting}, we use a 4-layer CNN on FMNIST, Cifar100, and Tiny-ImageNet. To evaluate \Method on a backbone larger than the 4-layer CNN, we also run ResNet-18~\cite{he2016deep} on Tiny-ImageNet. On AG News and Amazon Review, we use the fastText~\cite{joulinetal2017bag} and the 3-layer MLP~\cite{peng2019moment} as the backbones, respectively. Following previous work~\cite{zeng2014convolutional}, we use a HAR-CNN on HAR to process the sensor signal. 
As for the local learning rate $\eta$, we set $\eta=0.005$ for 4-layer CNN and 3-layer MLP, set $\eta=0.1$ for ResNet-18 and fastText, and set $\eta=0.01$ for HAR-CNN. 

\noindent\textbf{Statistically heterogeneous settings. \ } With the above six datasets, we create three popular statistically heterogeneous settings to simulate the FL environment: \textit{label skew}~\cite{mcmahan2017communication, NEURIPS2020_18df51b9, li2022federated}, \textit{feature shift}~\cite{li2020fedbn}, and \textit{real world}~\cite{zhang2022federated, gao2020end} settings. Specifically, we have two \textit{label skew} settings: the pathological setting~\cite{mcmahan2017communication, pmlrv139shamsian21a} and practical setting~\cite{NEURIPS2020_18df51b9, li2021model}. For the pathological \textit{label skew} setting, we sample data with label amount 2/10/20 for each client on FMNIST/Cifar100/Tiny-ImageNet from a total of 10/100/200 categories, with disjoint data and different numbers of data samples. For the practical \textit{label skew} setting, we sample data from FMNIST, Cifar100, Tiny-ImageNet, and AG News through the Dirichlet  distribution~\cite{NEURIPS2020_18df51b9}, denoted by $Dir(\beta)$. Concretely, we sample $q_{c, i} \sim Dir(\beta)$ ($\beta=0.1$ / $\beta=1$ by default for CV/NLP tasks~\cite{NEURIPS2020_564127c0}) and allocate a $q_{c, i}$ proportion of the samples with label $c$ to client $i$. For the \textit{feature shift} setting, following existing methods~\cite{feng2021kd3a, li2020fedbn}, we create four clients, each containing data from one domain on Amazon Review. For the \textit{real world} setting, the sensor signal data on HAR is naturally collected and stored on 30 clients with six activities. 

\noindent\textbf{Implementation Details. \ } Following pFedMe and FedRoD, unless explicitly specified, we have 20 clients with a client joining ratio $\rho=1$. On each client, we consider 75\% data as the training dataset and use the remaining 25\% data for evaluation. Following pFedMe, we report the best performance of the global model for traditional FL and the best average performance across personalized models for pFL. 
By default, we set the batch size to 10 and the number of local epochs to 1. We run 2000 iterations with three trials for all the methods on each task and report the mean and standard deviation. For more details and experimental results (\eg, the results in the \textit{feature shift} setting with different statistical heterogeneity, \ie, different $\beta$), please refer to supplementary materials.

\begin{table*}[ht]
  \centering
  \caption{The test accuracy (\%) on the CV and NLP tasks in \textit{label skew} settings.}
  \resizebox{\linewidth}{!}{
    \begin{tabular}{l|ccc|cccccc}
    \toprule
    \textbf{Settings} & \multicolumn{3}{c|}{\textbf{Pathological \textit{Label Skew} Setting}} & \multicolumn{5}{c}{\textbf{Practical \textit{Label Skew} Setting}} \\
    \midrule
     & FMNIST & Cifar100 & TINY & FMNIST & Cifar100 & TINY & TINY* & AG News\\
    \midrule
    \textbf{FedAvg} & 80.41$\pm$0.08 & 25.98$\pm$0.13 & 14.20$\pm$0.47 & 85.85$\pm$0.19 & 31.89$\pm$0.47 & 19.46$\pm$0.20 & 19.45$\pm$0.13 & 87.12$\pm$0.19 \\
    \textbf{FedProx} & 78.08$\pm$0.15 & 25.94$\pm$0.16 & 13.85$\pm$0.25 & 85.63$\pm$0.57 & 31.99$\pm$0.41 & 19.37$\pm$0.22 & 19.27$\pm$0.23 & 87.21$\pm$0.13 \\
    \midrule
    \textbf{Per-FedAvg} & 99.18$\pm$0.08 & 56.80$\pm$0.26 & 28.06$\pm$0.40 & 95.10$\pm$0.10 & 44.28$\pm$0.33 & 25.07$\pm$0.07 & 21.81$\pm$0.54 & 87.08$\pm$0.26 \\
    \textbf{pFedMe} & 99.35$\pm$0.14 & 58.20$\pm$0.14 & 27.71$\pm$0.40 & 97.25$\pm$0.17 & 47.34$\pm$0.46 & 26.93$\pm$0.19 & 33.44$\pm$0.33 & 87.08$\pm$0.18 \\
    \textbf{Ditto} & 99.44$\pm$0.06 & 67.23$\pm$0.07 & 39.90$\pm$0.42 & 97.47$\pm$0.04 & 52.87$\pm$0.64 & 32.15$\pm$0.04 & 35.92$\pm$0.43 & 91.89$\pm$0.17 \\
    \textbf{FedPer} & 99.47$\pm$0.03 & 63.53$\pm$0.21 & 39.80$\pm$0.39 & 97.44$\pm$0.06 & 49.63$\pm$0.54 & 33.84$\pm$0.34 & 38.45$\pm$0.85 & 91.85$\pm$0.24 \\
    \textbf{FedRep} & 99.56$\pm$0.03 & 67.56$\pm$0.31 & 40.85$\pm$0.37 & 97.56$\pm$0.04 & 52.39$\pm$0.35 & 37.27$\pm$0.20 & 39.95$\pm$0.61 & 92.25$\pm$0.20 \\
    \textbf{FedRoD} & 99.52$\pm$0.05 & 62.30$\pm$0.02 & 37.95$\pm$0.22 & 97.52$\pm$0.04 & 50.94$\pm$0.11 & 36.43$\pm$0.05 & 37.99$\pm$0.26 & 92.16$\pm$0.12 \\
    \textbf{FedPHP} & 99.30$\pm$0.13 & 63.09$\pm$0.04 & 37.06$\pm$0.57 & 97.38$\pm$0.16 & 50.52$\pm$0.16 & 35.69$\pm$3.26 & 29.90$\pm$0.51 & 90.52$\pm$0.19 \\
    \textbf{FedProto} & 99.49$\pm$0.04 & 69.18$\pm$0.03 & 36.78$\pm$0.07 & 97.40$\pm$0.02 & 52.70$\pm$0.33 & 31.21$\pm$0.16 & 26.38$\pm$0.40 & 96.34$\pm$0.58 \\
    \midrule
    \textbf{\Method} & \textbf{99.85$\pm$0.08} & \textbf{71.78$\pm$0.26} & \textbf{44.58$\pm$0.06} & \textbf{97.81$\pm$0.09} & \textbf{61.86$\pm$0.31} & \textbf{43.37$\pm$0.53} & \textbf{43.70$\pm$0.44} & \textbf{97.97$\pm$0.14} \\
    \bottomrule
    \end{tabular}}
  \label{tab:pathological}
\end{table*}

\subsection{Learned Features}
\label{sec:feat}

\begin{figure}[t]
	\centering
	\hfill
	\begin{subfigure}{0.29\linewidth}
	    \includegraphics[width=\linewidth]{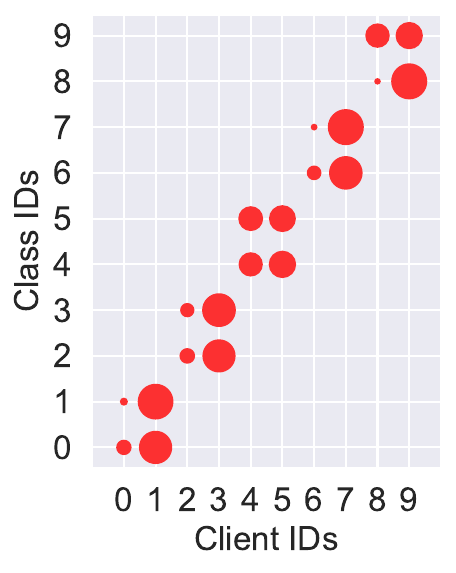}
	    \caption{Data distribution}
	    \label{fig:tsne_4}
	\end{subfigure}
	\hfill
	\begin{subfigure}{0.69\linewidth}
\includegraphics[width=\linewidth]{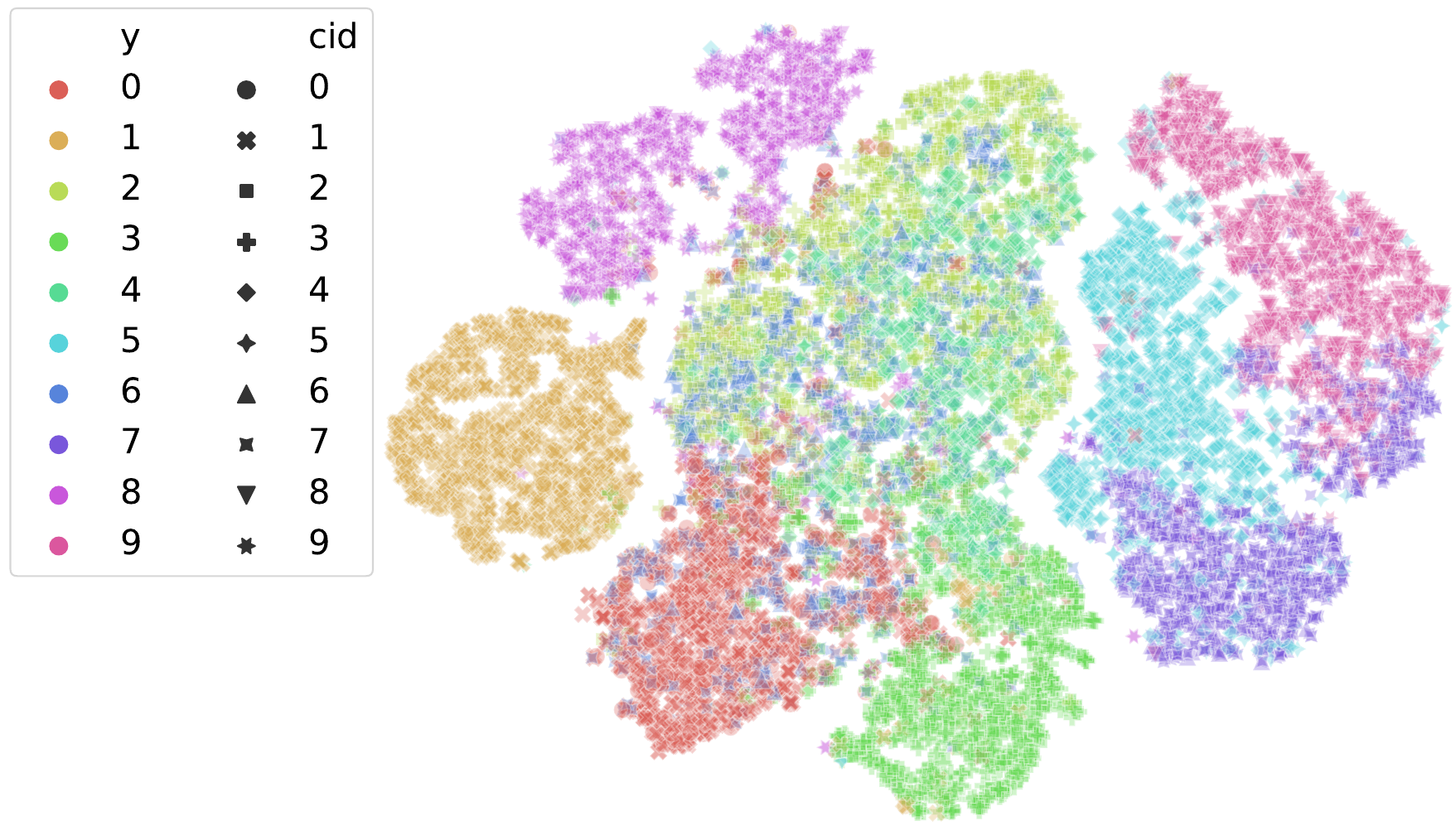}
	    \caption{FedPer}
	    \label{fig:tsne_1}
	\end{subfigure}
	\hfill
	\par\medskip
	\hfill
	\begin{subfigure}{0.49\linewidth}
	    \includegraphics[width=\linewidth]{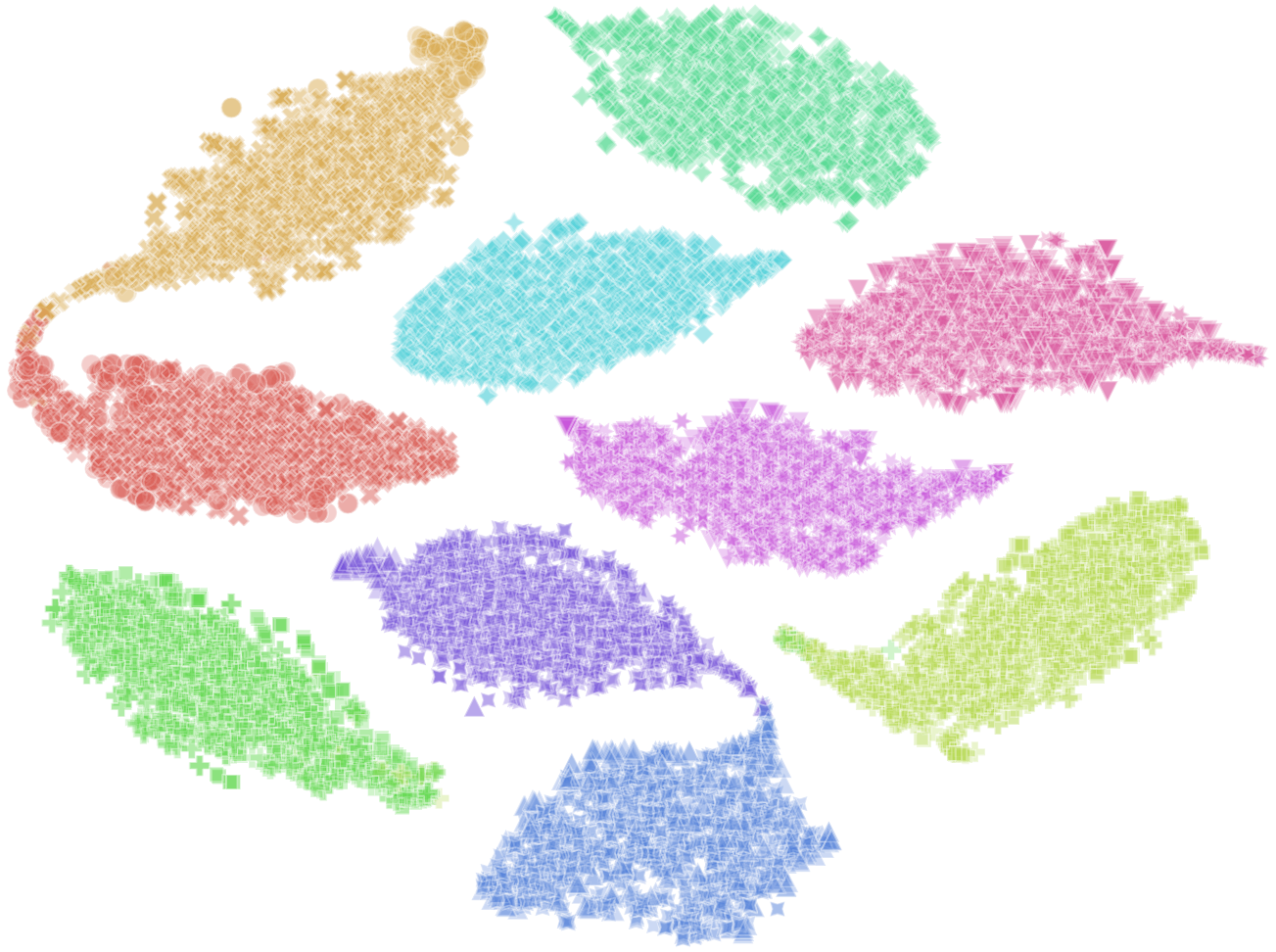}
	    \caption{FedProto}
	    \label{fig:tsne_2}
	\end{subfigure}
	\hfill
	\begin{subfigure}{0.49\linewidth}
\includegraphics[width=\linewidth]{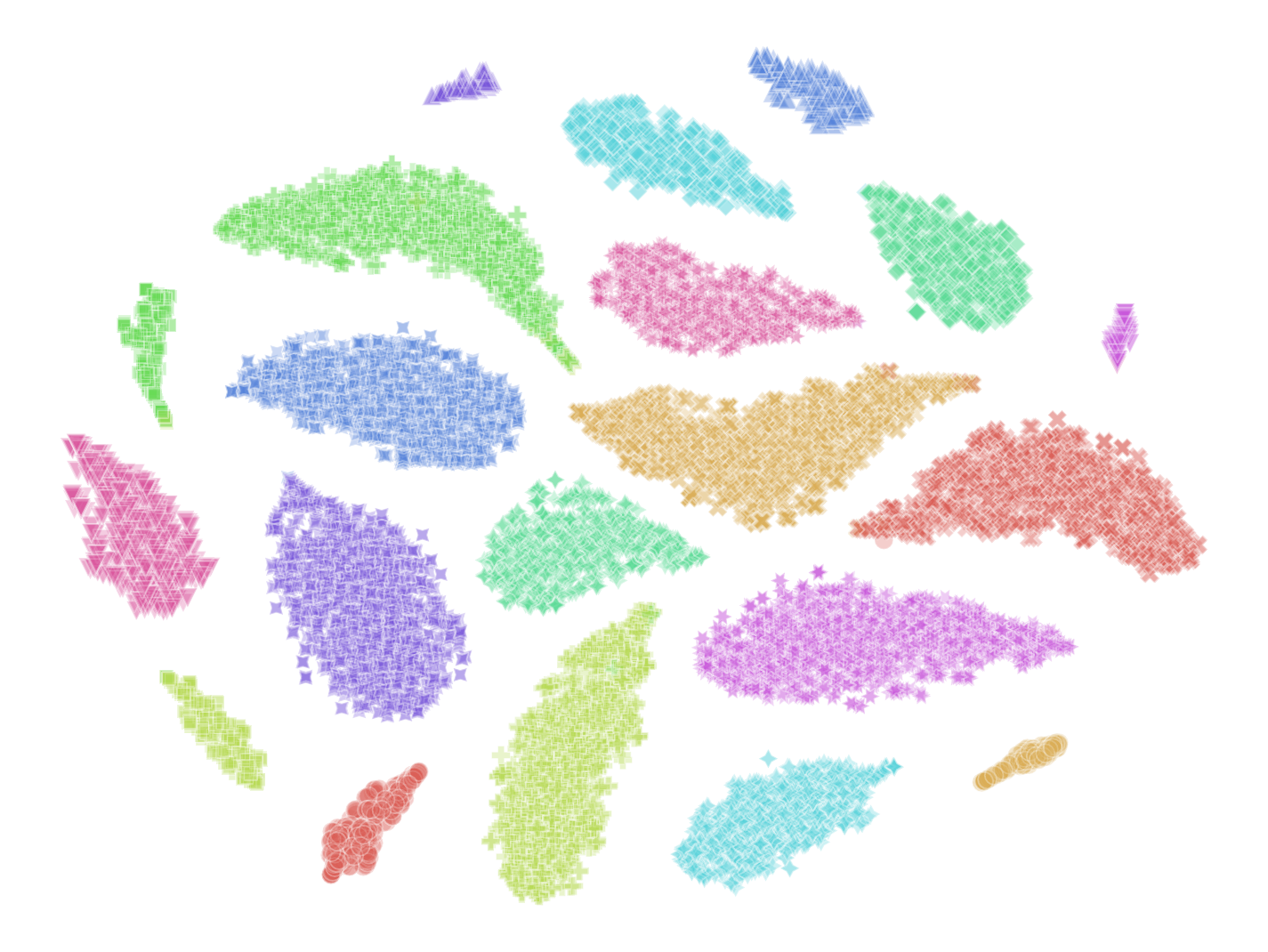}
	    \caption{\Method}
	    \label{fig:tsne_3}
	\end{subfigure}
	\hfill
\setlength{\belowcaptionskip}{-10pt}
	\caption{(a): Data distribution on each client; the size of red circles means the number of samples. (b), (c), and (d): t-SNE visualizations of feature vectors on FMNIST with 10 clients; ``cid'' means client ID. \textit{Best viewed in color.} }
	\label{fig:tsne}
\end{figure}

Here, for easy visualization, we experiment on FMNIST in the pathological \textit{label skew} setting using ten clients and keep other settings constant. As shown in \Cref{fig:tsne}, we use t-SNE~\cite{van2008visualizing} to visualize the feature vectors extracted by FedPer, FedProto, and our \Method. 

The classification boundary is not discriminative in FedPer and the boundaries in FedProto intersect, \eg, the boundaries of category 0 and 1 in \Cref{fig:tsne_2}, while \Cref{fig:tsne_3} has no intersection. By simultaneously considering global guidance and personalized tasks, \Method can learn discriminative features and distinguish the data distribution of each client, revealing excellent personalization performance of \Method in feature extraction. 

\subsection{Effectiveness}
\label{sec:eff}

Then, we compare our \Method with baselines under the \textit{label skew}, \textit{feature shift}, and \textit{real world} settings. Due to the limited space, we use ``TINY'' and ``TINY*'' to represent using the 4-layer CNN and ResNet-18 (trained from scratch) on Tiny-ImageNet, respectively. 

\noindent\textbf{\textit{Label skew} settings. \ } We show the results regarding the \textit{label skew} settings in \Cref{tab:pathological}. \Method achieves superior performance in both pathological and practical settings. Concretely, in the practical setting on Cifar100, \Method outperforms the best baseline Ditto by \textbf{8.99\%} with only 0.31\% standard deviation. FedRep performs well on Tiny-ImageNet in the pathological setting, but \Method outperforms it by 6.10\% in the practical setting. Since there are many embeddings of tokens in NLP tasks, the feature guidance per category in \Method and FedProto is beneficial for embedding learning in the \textit{label skew} setting. \Method achieves 1.63\% improvement over FedProto and outperforms other baselines by 5.72\% on AG News. 

Next, we point out why \Method outperforms other baselines with the experimental results.
(1) \textbf{\Method v.s. FedPer \& FedRep \& FedRoD}: On each client, the feature extractor trained in FedPer and FedRep only learns personalized feature information, while the feature extractor in FedRoD only learns the global one. Unlike them, \Method locally learns both the global and personalized feature information, so \Method outperforms FedPer/FedRep/FedRoD by 12.23\%/9.47\%/10.92\% on Cifar100 in the practical setting. 
(2) \textbf{\Method v.s. FedProto \& FedPHP}: FedPHP/FedProto guides feature extraction with global features/prototypes throughout the FL process, but the large backbone ResNet-18 cannot learn to extract features well in early iterations. Then, poor global features and prototypes mislead local training, so FedPHP and FedProto achieve low accuracy on the large backbone ResNet-18. Besides, due to the classification boundary intersection, FedProto performs worse on the dataset with more categories (\eg, Tiny-ImageNet with 200 categories). By sharing the feature extractor and guiding the feature extraction with the trainable global embeddings, \Method outperforms FedProto by 17.32\% on TINY*.

\begin{figure}[ht]
	\centering
	\begin{subfigure}{\linewidth}
    	\includegraphics[width=\linewidth]{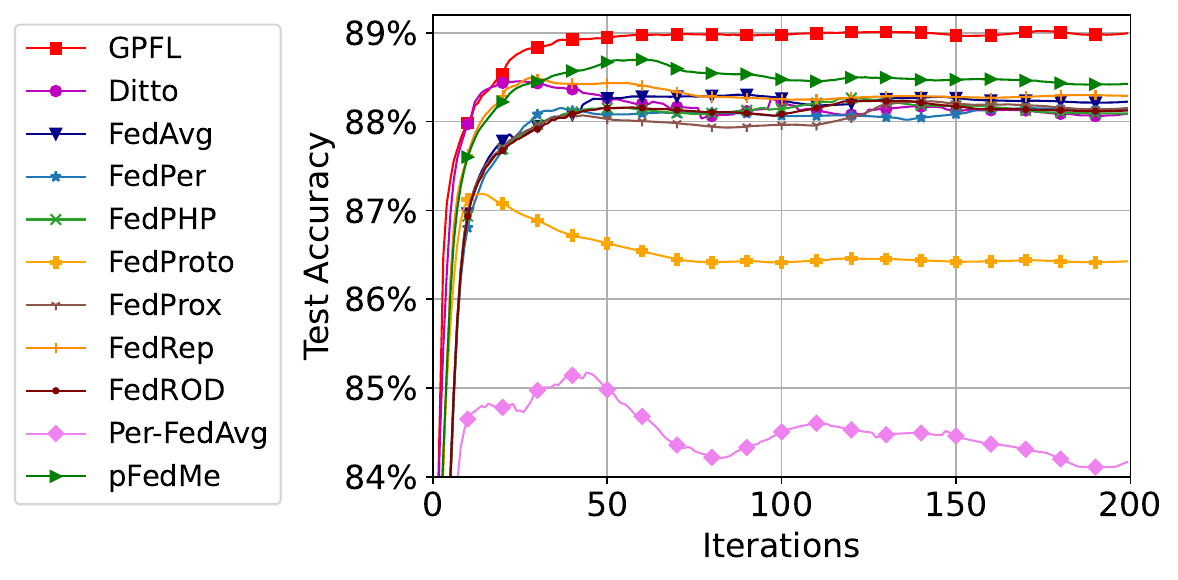}
    	\caption{Test accuracy curves in the \textit{feature shift} setting.}
	\label{fig:amazon}
	\end{subfigure}
	
	\begin{subfigure}{\linewidth}
    	\includegraphics[width=\linewidth]{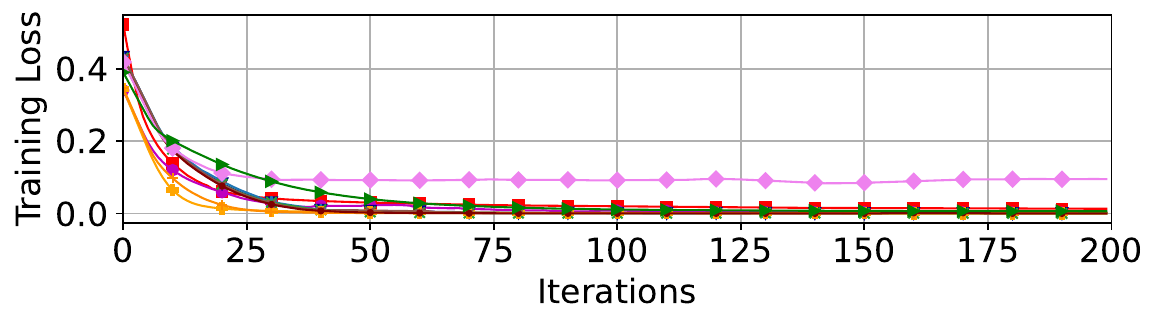}
    	\caption{Training loss curves in the \textit{feature shift} setting.}
	\label{fig:amazon_loss}
	\end{subfigure}
	\caption{Curves (smoothed) on Amazon Review dataset.}
\end{figure}

\noindent\textbf{\textit{Feature shift} setting. \ } Using the Amazon Review dataset, each client contains the data from one domain with the personalized task of classifying the samples into positive or negative emotions. In other words, the data on every client belong to two categories, which is different from the \textit{label skew} settings. 
In \Cref{fig:amazon}, we find that the traditional FL methods FedAvg and FedProx achieve good performance with a little gap compared to the pFL methods. 
It means that traditional FL methods suffer more in the \textit{label skew} setting than in the \textit{feature shift} setting. 
Besides, FedAvg can maintain its performance after reaching the best accuracy. In contrast, the accuracy curves of several pFL methods, including Ditto, pFedMe, FedRep, FedProto, and Per-FedAvg, exhibit a drop in accuracy after reaching the peak accuracy, despite their training having converged (\Cref{fig:amazon_loss}), which means overfitting. FedProto and Per-FedAvg achieve poor performance with a significant drop. On the contrary, \Method performs the best and maintains its performance when converged, as the global information in ${\rm \compo}$ mitigates the overfitting of personalized models. 
FedRoD does not show superiority here, as its BSM loss is only designed for the \textit{label skew} settings to tackle statistical heterogeneity.

\begin{table*}[ht]
  \centering
  \caption{The test accuracy (\%) on the IoT task regarding effectiveness and the CV task regarding scalability.}
  \resizebox{!}{!}{
    \begin{tabular}{l|c|cccc|cc}
    \toprule
    \textbf{Tasks} & \textbf{Effectiveness} & \multicolumn{6}{c}{\textbf{Scalability}}\\
    \midrule
    \textbf{Clients} & $N=30$ & $N=30$ & $N=50$ & $N=100$ & $N=500$ & $N=10|50$ & $N=30|50$ \\
    \midrule
    \textbf{FedAvg} & 87.20$\pm$0.27 & 31.15$\pm$0.05 & 31.90$\pm$0.27 & 31.95$\pm$0.37 & 29.51$\pm$0.73 & 25.28$\pm$0.32 & 29.04$\pm$0.21 \\
    \textbf{FedProx} & 88.34$\pm$0.24 & 31.21$\pm$0.08 & 31.94$\pm$0.30 & 31.97$\pm$0.24 & 29.84$\pm$0.81 & 25.65$\pm$0.34 & 29.04$\pm$0.36 \\
    \midrule
    \textbf{Per-FedAvg} & 77.12$\pm$0.17 & 41.57$\pm$0.21 & 44.31$\pm$0.20 & 36.07$\pm$0.24 & / & 40.20$\pm$0.21 & 42.96$\pm$0.42 \\
    \textbf{pFedMe} & 91.57$\pm$0.12 & 47.04$\pm$0.28 & 48.36$\pm$0.64 & 46.45$\pm$0.18 & 31.30$\pm$0.89 & 40.27$\pm$0.54 & 42.19$\pm$0.38 \\
    \textbf{Ditto} & 91.53$\pm$0.09 & 52.53$\pm$0.42 & 54.22$\pm$0.04 & 52.89$\pm$0.22 & 30.24$\pm$0.72 & 48.23$\pm$0.35 & 50.98$\pm$0.29 \\
    \textbf{FedPer} & 75.58$\pm$0.13 & 44.98$\pm$0.20 & 44.22$\pm$0.18 & 40.37$\pm$0.41 & 30.56$\pm$0.59 & 43.64$\pm$0.42 & 43.54$\pm$0.43 \\
    \textbf{FedRep} & 80.44$\pm$0.42 & 50.24$\pm$0.01 & 47.41$\pm$0.18 & 44.61$\pm$0.20 & 31.92$\pm$0.71 & 46.85$\pm$0.12 & 47.63$\pm$0.26 \\
    \textbf{FedRoD} & 89.91$\pm$0.23 & 50.11$\pm$0.03 & 49.38$\pm$0.01 & 46.65$\pm$0.22 & 34.61$\pm$0.98 & 46.32$\pm$0.02 & 49.15$\pm$0.12 \\
    \textbf{FedPHP} & 87.94$\pm$0.54 & 49.28$\pm$0.06 & 52.44$\pm$0.16 & 49.70$\pm$0.31 & 30.26$\pm$0.84 & 45.71$\pm$0.21 & 48.65$\pm$0.24 \\
    \textbf{FedProto} & 84.73$\pm$0.32 & 52.32$\pm$0.18 & 50.29$\pm$0.18 & 47.11$\pm$0.15 & 34.91$\pm$0.83 & 49.23$\pm$0.28 & 50.49$\pm$0.32 \\
    \midrule
    \textbf{\Method} & \textbf{93.76$\pm$0.16} & \textbf{60.96$\pm$0.65} & \textbf{60.98$\pm$0.32} & \textbf{57.11$\pm$0.21} & \textbf{37.28$\pm$0.63} & \textbf{53.26$\pm$0.39} & \textbf{59.18$\pm$0.53} \\
    \bottomrule
    \end{tabular}}
  \label{tab:beta}
\end{table*}

\noindent\textbf{\textit{Real world} setting. \ } From \Cref{tab:beta}, we find that the regularization-based pFL methods (pFedMe and Ditto) perform well on HAR and the regularization-based traditional FL method FedProx outperforms FedAvg by 1.14\%. 
However, most other pFL methods, \eg, Per-FedAvg, FedPer, FedRep, FedPHP, and FedProto perform worse than traditional FL methods (FedAvg and FedProx). FedAvg outperforms Per-FedAvg and FedPer by 10.08\% and 11.62\%, respectively. The ${\rm \compo}$ stores the global embedding of each activity, which guides the feature extractor to extract the global characteristic of one kind of human activity. Meanwhile, the personalized task can also guide the feature extractor to extract user-specific characteristics. Thus, our \Method outperforms all baselines by up to 2.19\% in this scenario. 

\subsection{Scalability}

Based on MOON~\cite{li2021model}, we split the Cifar100 dataset into 20/30/50/100/500 sub-datasets to form 20/30/50/100/500 clients, respectively, in the practical \textit{label skew} settings ($\beta=0.1$). We have shown the results for 20 clients in \Cref{tab:pathological}, so we only show the results for 30/50/100/500 clients in \Cref{tab:beta}. The  meta-learning in Per-FedAvg requires at least two batches of data, which is invalid on some clients in our unbalanced settings when $N=500$. Since the total sample amount is constant on Cifar100, the average sample amount per client decreases when $N$ increases and it is unreasonable to compare results between different $N$. 

In \Cref{tab:beta}, \Method still outperforms other baselines with various $N$. When $N=500$, it is hard to achieve personalization without enough client data for pFL methods. Many pFL baselines, \eg, Ditto, FedPer, and FedPHP perform similarly with FedAvg in this situation. Since ${\rm \compo}$ provides extra information and relieves the data shortage issue on each client, \Method outperforms all baselines. 

In practice, clients join the FL process with their data, and more clients means more total data in FL. To simulate a more realistic scenario, we randomly select 10/30/50 clients from the 50 clients generated above to form a setting with a total of 10/30/50 clients, denoted by $N=10|50$ / $N=30|50$ / $N=50$. In these settings, as shown in \Cref{tab:beta}, all the methods benefit from the client amount increase, and \Method still outperforms other baselines, showing the scalability of \Method in practice.

\subsection{Fairness}

\begin{table}[ht]
  \centering
  \caption{The fairness, \ie, standard deviation (\%, $\downarrow$) [the coefficient of variation ($\times 10^{-2}$, $\downarrow$)] of the local accuracy on clients when reaching the best test accuracy on the CV, NLP, and IoT tasks in pathological \textit{label skew} (paLS), practical \textit{label skew} (prLS, $\beta=0.1$), \textit{feature shift}, and \textit{real world} settings. }
  \resizebox{\linewidth}{!}{
    \begin{tabular}{l|cccc}
    \toprule
    \textbf{Settings} & \textbf{paLS} & \textbf{prLS} & \textbf{\textit{Feature Shift}} & \textbf{\textit{Real World}} \\
    \midrule
    \textbf{Clients} & $N=20$ & $N=100$ & $N=4$ & $N=30$ \\
    \midrule
    \textbf{Datasets} & TINY & Cifar100 & Amazon Review & HAR \\
    \midrule
    \textbf{FedAvg} & 3.57 [25.14] & 7.06 [22.10] & 1.62 [1.84] & 17.10 [19.61] \\
    \textbf{FedProx} & 3.51 [25.34] & 7.08 [22.15] & 1.60 [1.81] & 17.35 [19.64] \\
    \midrule
    \textbf{Per-FedAvg} & 3.27 [11.65] & 8.13 [22.54] & 2.82 [3.29] & 14.15 [18.35] \\
    \textbf{pFedMe} & 3.36 [12.12] & 8.19 [17.63] & 1.99 [2.25] & 12.65 [13.81] \\
    \textbf{Ditto} & 3.84 [9.62] & 9.89 [18.70] & 2.12 [2.40] & 13.20 [14.42] \\
    \textbf{FedPer} & 3.39 [8.51] & 8.91 [22.07] & 2.18 [2.47] & 19.49 [25.79] \\
    \textbf{FedRep} & 3.53 [8.64] & 8.99 [20.15] & 2.15 [2.43] & 21.16 [26.30] \\
    \textbf{FedRoD} & 3.46 [9.12] & 8.87 [19.01] & 2.24 [2.54] & 16.93 [18.83] \\
    \textbf{FedPHP} & 3.81 [10.28] & 9.45 [19.01] & 1.96 [2.22] & 13.81 [15.70] \\
    \textbf{FedProto} & 4.13 [11.23] & 9.98 [21.18] & 1.82 [2.08] & 11.77 [13.89] \\
    \midrule
    \textbf{\Method} & \textbf{3.21 [7.20]} & \textbf{8.05 [14.10]} & \textbf{1.62 [1.80]} & \textbf{8.42 [8.98]} \\
    \bottomrule
    \end{tabular}}
  \label{tab:fairness}
\end{table}

According to the work~\cite{pillutla2022federated}, personalization may result in poorer performance on some devices despite improving the average. It is essential to improve both performance and fairness when designing a pFL method. Here, following Ditto~\cite{li2021ditto}, we evaluate the fairness of all the methods through the standard deviation of the local accuracy on clients when reaching the best-averaged accuracy (\ie, the test accuracy mentioned above), as shown in \Cref{tab:fairness}. To weaken the effect of the test accuracy magnitude for a more fair comparison, we also follow the work~\cite{jain1984quantitative} to use the coefficient of variation metric for fairness evaluation. 

In \Cref{tab:fairness}, our \Method outperforms other pFL methods, especially in the \textit{real world} setting and the settings with many clients, because sharing global information among clients promotes fairness. By learning both the global and personalized feature information, \Method achieves a higher accuracy with lower discrimination compared to FedPer, FedRep, and FedRoD, which focus on learning only one kind of feature information during local training. The traditional FL methods achieve a low standard deviation but a high coefficient of variation, as their personalization performance is poor. Clients in FedProto only share prototypes but keep the model parameters secret, which limits the capacity of global information and leads to low fairness.

\subsection{Stability}

\begin{table}[ht]
  \centering
  \caption{The test accuracy (\%) of the pFL methods on Cifar100 with $N=50$, $\beta=0.1$, and $\rho \le 1$.}
  \resizebox{\linewidth}{!}{
    \begin{tabular}{l|ccc}
    \toprule
    & $\rho=1$ & $\rho \in [0.5, 1]$ & $\rho \in [0.1, 1]$ \\
    \midrule
    \textbf{Per-FedAvg} & 44.31$\pm$0.20 & 43.66$\pm$1.38 & 43.63$\pm$1.07\\
    \textbf{pFedMe} & 48.36$\pm$0.64 & 43.28$\pm$0.85 & 41.71$\pm$1.02\\
    \textbf{Ditto} & 50.59$\pm$0.22 & 49.78$\pm$0.36 & 48.33$\pm$3.27\\
    \textbf{FedPer} & 44.22$\pm$0.18 & 44.12$\pm$0.21 & 44.07$\pm$0.27\\
    \textbf{FedRep} & 47.41$\pm$0.18 & 46.93$\pm$0.21 & 46.61$\pm$0.22\\
    \textbf{FedRoD} & 49.38$\pm$0.01 & 49.07$\pm$0.43 & 47.80$\pm$1.35\\
    \textbf{FedPHP} & 50.23$\pm$0.12 & 45.19$\pm$0.07 & 44.43$\pm$0.12\\
    \textbf{FedProto} & 50.29$\pm$0.18 & 49.45$\pm$0.21 & 46.05$\pm$4.03 \\
    \midrule
    \textbf{\Method} & \textbf{60.98$\pm$0.32} & \textbf{60.60$\pm$0.51} & \textbf{60.04$\pm$0.28} \\
    \bottomrule
    \end{tabular}}
  \label{tab:dynamic}
\end{table}

In real world scenarios, some clients cannot join the whole FL process because of low battery, lack of computation resources, unstable network, \etc. Here, we simulate this scenario by varying the client joining ratio $\rho$ in every iteration on the Cifar100 dataset. Specifically, we uniformly sample a value for $\rho$ within the given range in each iteration. 

In \Cref{tab:dynamic}, our \Method still maintains its superiority among the pFL methods, while some baselines perform worse with a more extensive range of $\rho$. For example, pFedMe and FedPHP drop 6.65\% and 5.80\% from $\rho=1$ to $\rho \in [0.1, 1]$, respectively. Furthermore, Per-FedAvg, pFedMe, Ditto, FedRoD, and FedProto achieve erratic performance (large standard deviations) in dynamic settings. 

\subsection{Privacy}
\label{sec:privacy}

\begin{table}[ht]
  \centering
  \caption{PSNR (dB, $\downarrow$) values for privacy evaluation on Cifar100 in \textit{label skew} setting. ``Fed'' is omitted for method names. }
  \resizebox{\linewidth}{!}{
    \begin{tabular}{c|c|cc|cc|ccc}
    \toprule
    \textbf{Per-\textit{fe}} & \textbf{Rep-\textit{fe}} & \textbf{Avg-\textit{fe}} & \textbf{Avg} & \textbf{RoD-\textit{fe}} & \textbf{RoD} & \textbf{\Method-\textit{fe}} & \textbf{\Method-\textit{sfe}}& \textbf{\Method-\textit{sm}}  \\
    \midrule
    7.94 & 7.73 & 6.92 & 7.30 & 6.82 & 7.48 & 6.56 & \textbf{6.41} & 6.71 \\
    \bottomrule
    \end{tabular}}
  \label{tab:privacy}
\end{table}

Following FedCG, we evaluate the privacy-preserving ability of \Method with representative baselines in \textit{Peak Signal-to-Noise Ratio} (PSNR)~\cite{hore2010image}. For FedPer, FedRep, FedAvg, and FedRoD, we conduct DLG attack for \underline{f}eature \underline{e}xtractor (with suffix ``-\textit{fe}''). For FedAvg and FedRoD, we also conduct DLG attacks for the entire model. For \Method, we conduct DLG attack for feature extractor, p\underline{s}eudo \underline{f}eature \underline{e}xtractor (with suffix ``-\textit{sfe}''), and p\underline{s}eudo \underline{m}odel (with suffix ``-\textit{sm}''). 

In \Cref{tab:privacy}, all the PSNR values that correspond to \Method are smaller than other baselines, which supports the analysis in \cref{sec:privacy_ana}. Since the parameters of the last FC layer in a backbone also contain class information~\cite{niu2022federated}, the entire model is more susceptible to privacy leakage than the feature extractor in FedAvg and FedRoD. However, due to the global characteristic of ${\rm \compo}$, \Method-\textit{sm} still maintains the privacy-preserving ability.

\section{Ablation Study}
\label{sec:ablation}

\begin{figure}[ht]
	\centering
	\hfill
	\begin{subfigure}{0.57\linewidth}
	    \includegraphics[width=\linewidth]{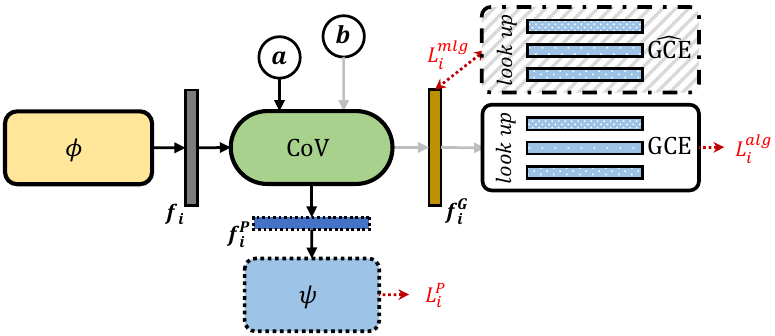}
	    \caption{\textit{w/o ${\rm PCI}$}}
	    \label{fig:abl_1}
	\end{subfigure}
	\hfill
	\begin{subfigure}{0.41\linewidth}
	    \includegraphics[width=\linewidth]{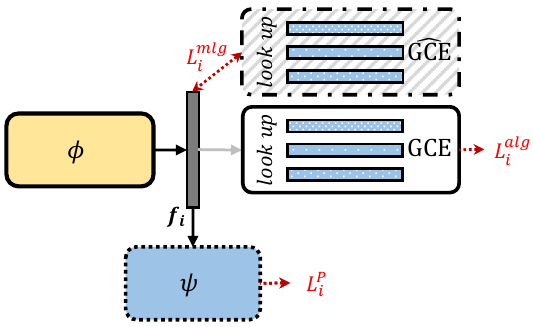}
	    \caption{\textit{w/o ${\rm \Key}$}}
	    \label{fig:abl_2}
	\end{subfigure}
	\par\bigskip
	\hfill
	\begin{subfigure}{0.58\linewidth}
	    \includegraphics[width=\linewidth]{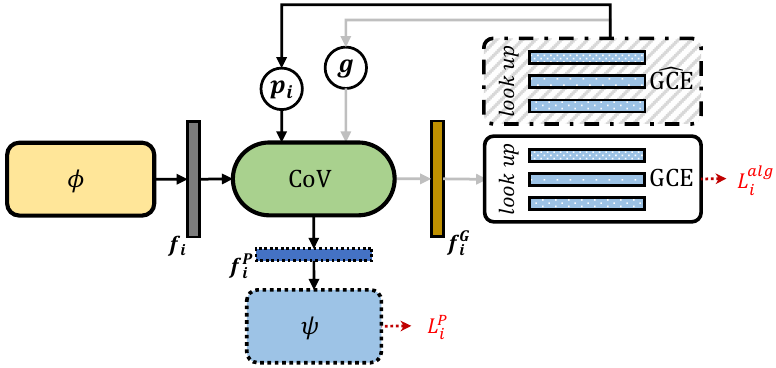}
	    \caption{\textit{w/o $\mathcal{L}_i^{mlg}$}}
	    \label{fig:abl_3}
	\end{subfigure}
	\hfill
	\begin{subfigure}{0.35\linewidth}
	    \includegraphics[width=\linewidth]{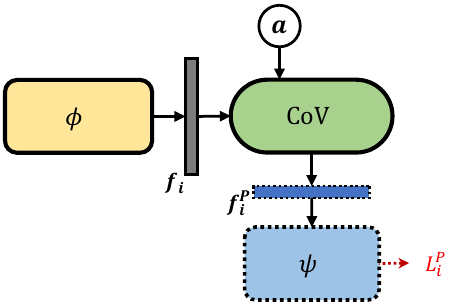}
	    \caption{\textit{w/o ${\rm \compo}$}}
	    \label{fig:abl_4}
	\end{subfigure}
	\hfill
\setlength{\belowcaptionskip}{-10pt}
	\caption{Illustration of variants for ablation study.}
	\label{fig:abl}
\end{figure}

To further show the effectiveness of each proposed module, we remove them from \Method and create five variants (``\textit{w/o}'' is short for ``without''): (a) \textit{w/o} personalized conditional input (${\rm PCI}$), (b) \textit{w/o} ${\rm \Key}$, (c) \textit{w/o} $\mathcal{L}_i^{mlg}$, (d) \textit{w/o} ${\rm \compo}$ ($\widehat{\rm \compo}$ also disappears), and (e) \textit{w/o} ${\rm \Key}$ \textit{\&} ${\rm \compo}$ (FedPer). As shown in \Cref{fig:abl}, we input the constant vector ${\bm a}$ and ${\bm b}$ instead of the dynamic ${\bm p}_i$ and ${\bm g}$ to remove ${\rm PCI}$. 
We report the test accuracy of \Method and its five variants in \Cref{tab:abl}. 

\begin{table}[ht]
  \centering
  \caption{The accuracy (\%) of \Method and its variants on TINY*. }
  \resizebox{\linewidth}{!}{
    \begin{tabular}{c|ccccc}
    \toprule
    \textbf{\Method} & \textbf{\textit{w/o} ${\rm PCI}$} & \textbf{\textit{w/o} ${\rm \Key}$} & \textbf{\textit{w/o} $\mathcal{L}_i^{mlg}$} & \textbf{\textit{w/o} ${\rm \compo}$} & \textbf{FedPer} \\
    \midrule
    \textbf{43.70} & 42.74 & 40.23 & 41.72 & 39.48 & 38.45\\
    \bottomrule
    \end{tabular}}
  \label{tab:abl}
\end{table}

We analyze each module according to \Cref{tab:abl}. (a) Like ${\bm g}$, ${\bm a}$ and ${\bm b}$ are identical across clients, so \Method inputs local data distribution information for ${\rm \Key}$ while \textit{w/o} ${\rm PCI}$ does not. With this personalized information, \Method performs better than \textit{w/o} ${\rm PCI}$. Even with the identical ${\bm a}$/${\bm b}$, \textit{w/o} ${\rm PCI}$ still performs well thanks to end-to-end training. (b) Removing ${\rm \Key}$ causes a 3.47\% accuracy decrease, as guiding one feature vector to learn both global and personalized information simultaneously is confusing. 
(c) The accuracy gap between \Method and \textit{w/o} $\mathcal{L}_i^{mlg}$ shows the effectiveness of the magnitude-level global guidance, since \textit{w/o} $\mathcal{L}_i^{mlg}$ only removes the $\mathcal{L}_i^{mlg}$ objective. (d) The accuracy decreases further when we remove the angle-level global guidance. Without learning global information during local training, the accuracy drops by 4.22\%. As the trainable affine mapping adaptively adjusts the original features, \textit{w/o} ${\rm \compo}$ still slightly outperforms FedPer. 




\section{Conclusion}

For the collaborative learning and personalization goals of pFL, we propose \Method to simultaneously learn global and personalized information on the client. We show the superiority of \Method through extensive experiments
regarding effectiveness, scalability, fairness, stability, and privacy. 
\Method outperforms SOTA pFL methods by a large margin. 
Besides, we show the effectiveness of each proposed module. 

\section*{Acknowledgements}

This work was supported in part by the Shanghai Key Laboratory of Scalable Computing and Systems, National Key R\&D Program of China (2022YFB4402102), Internet of Things special subject program, China Institute of IoT (Wuxi), Wuxi IoT Innovation Promotion Center (2022SP-T13-C), Industry-university-research Cooperation Funding Project from the Eighth Research Institute in China Aerospace Science and Technology Corporation (Shanghai) (USCAST2022-17), and Intel Corporation (UFunding 12679). This work was also partially supported by the Program of Technology Innovation of the Science and Technology Commission of Shanghai Municipality (Granted No. 21511104700 and 22DZ1100103).


{\small
\bibliographystyle{ieee_fullname}
\bibliography{main}
}

\end{document}